\documentclass{article}
\usepackage{ijcai26}

\usepackage{times}
\usepackage{soul}
\usepackage{url}
\usepackage[hidelinks]{hyperref}
\usepackage[utf8]{inputenc}
\usepackage[small]{caption}
\usepackage{graphicx}
\usepackage{amsmath}
\usepackage{amsthm}
\usepackage{booktabs}
\usepackage{algorithm}
\usepackage{algorithmic}
\usepackage[switch]{lineno}
\newtheorem{definition}{Definition}[section]
\usepackage{subcaption}
\usepackage{amssymb}

\usepackage{hyperref}

\title{Towards Visually Explaining Statistical Tests with Applications in Biomedical Imaging}

\author{
Masoumeh Javanbakhat$^{1}$\and
Piotr Komorowski$^{2}$\and
Dilyara Bareeva$^{2}$\and
Wei-Chang Lai$^{1}$\and
Wojciech Samek$^{2}$\and
Christoph Lippert$^{1}$
\affiliations
$^1$Digital Health-Machine Learning group, Hasso-Plattner Institute, Potsdam, Germany\\
$^2$Department of Artificial Intelligence, Fraunhofer Heinrich Hertz Institute, Berlin, Germany
\emails
\{Masoumeh.Javanbakhat, Wei-Cheng.Lai, Christoph.Lippert\}@hpi.de,\\
\{piotr.komorowski, dilyara.bareeva, wojciech.samek\}@hhi.fraunhofer.de 
}

\begin{document}

\maketitle

\begin{abstract}
Deep neural two-sample tests have recently shown strong power for detecting distributional differences between groups, yet their black-box nature limits interpretability and practical adoption in biomedical analysis. 
Moreover, most existing post-hoc explainability methods rely on class labels, making them unsuitable for label-free statistical testing settings. We propose an explainable deep statistical testing framework that augments deep two-sample tests with sample-level and feature-level explanations, revealing which individual samples and which input features drive statistically significant group differences. Our method highlights which image regions and which individual samples contribute most to the detected group difference, providing spatial and instance-wise insight into the test’s decision.
Applied to biomedical imaging data, the proposed framework identifies influential samples and highlights anatomically meaningful regions associated with disease-related variation. This work bridges statistical inference and explainable AI, enabling interpretable, label-free population analysis in medical imaging.
\end{abstract}

\section{Introduction}
\label{sec:intro}
\subsection{Statistical Tests}
Two-sample tests are a fundamental tool in statistical inference for determining whether two sets of observations are drawn from the same underlying distribution (the null hypothesis) or not (the alternative hypothesis). Two-sample tests can be broadly divided into parametric and non-parametric methods. Parametric two-sample tests, such as the Student’s t-test, make strong assumptions about the distribution of the data (e.g., Gaussian). However, parametric tests may fail when their assumptions about the data distribution are invalid. Non-parametric tests, on the other hand, do not assume a specific parametric form for the underlying distribution and can thus be potentially applied in a wider range of application scenarios. Computing nonparametric test statistics, however, can be costly as it may require applying resampling schemes or computing higher-order statistics. 

To address this limitation, deep two-sample tests have been recently proposed, leveraging deep neural networks to learn data-driven representations in which distributional differences become more separable \cite{2sample-test}. By combining representation learning with classical hypothesis testing, these methods achieve high statistical power and can be applied to a wide range of high-dimensional non-tabular data modalities. Despite their effectiveness (high power), deep two-sample tests operate as black-box statistical procedures, providing no insight into which \textit{features} or \textit{individual samples} drive the rejection of the null hypothesis—a limitation that hinders their practical adoption in scientific and medical applications. 

\subsection{Post-Hoc Explainability}
The interpretability of black-box models has received growing attention in machine learning, leading to the development of \textit{explainable artificial intelligence} (XAI) \cite{Adi}. Post-hoc explainability methods aim to interpret trained models by identifying which features or samples influence their outputs.
\textbf{Feature-level explainability} focuses on identifying input features or regions that are most influential for a model’s prediction. Techniques such as Input$\times$Gradient \cite{Sailiency}, Integrated Gradients \cite{IG},  Layer-wise Relevance Propagation \cite{10.1371/journal.pone.0130140}, and perturbation-based methods \cite{lundberg2017unified} have become standard tools for interpreting deep models. These approaches fundamentally rely on the existence of class labels or decision boundaries, as attributions are computed with respect to a specific prediction target. \textbf{Sample-level explainability}, on the other hand, aims to identify influential training examples that contribute to a model’s output. Representative methods include Deep k-Nearest Neighbors \cite{KNN,DeepKNN}, influence functions \cite{Is,IS2}, TraceIn \cite{TraceInf}, and SimplEx \cite{SampleEx}. These techniques have not been applied to statistical testing, where identifying samples that drive test rejection remains unexplored.
\subsection{Motivation}
Despite their success, existing post-hoc explainability methods are inherently designed to explain predictive models, where explanations are conditioned on a model output such as a class label or a loss value. Even recent extensions to unsupervised or self-supervised learning focus on explaining learned representations or similarity relationships \cite{LFree}, rather than decisions arising from statistical inference. 

In the context of two-sample testing, interpretability raises fundamentally different questions. Rather than asking why a model predicts a certain label, one seeks to understand which \textit{features} or \textit{spatial regions} are responsible for the detected distributional difference between groups, and which \textit{samples} drive the rejection of the null hypothesis. Answering these questions is critical in scientific and medical applications, where statistical significance alone is insufficient and insights into underlying mechanisms are required. However, current XAI frameworks do not address these questions for deep statistical tests, leaving a critical gap between powerful testing procedures and interpretable population-level analysis.

In this work, we address this limitation by introducing an explainability framework that provides \textbf{sample-level} and \textbf{feature-level} insight into deep two-sample tests. By turning statistically significant group differences into interpretable, sample- and feature-level insights, our explainable deep statistical testing framework enables actionable analysis in biomedical imaging. This supports downstream tasks such as patient stratification, quality control, clinical trial enrichment, and biomarker discovery, by making population-level differences transparent, robust, and biologically meaningful.
\subsection{Related Works}
Kernel-based hypothesis testing, particularly the Maximum Mean Discrepancy (MMD), provides a principled and well-established framework for comparing distributions \cite{JMLR:v13:gretton12a}. Recent work has enhanced test power by learning data-driven representations \cite{2sample-test,pmlr-v119-liu20m}. Prior work has explored incorporating interpretability into test design. \citeauthor{jitkrittum2016interpretable} \shortcite{jitkrittum2016interpretable} proposed optimizing locations in spatial or frequency domains to maximize a lower bound on test power for a statistical test, revealing an interpretable indication of where two distributions differ. \citeauthor{lopez2016revisiting} \shortcite{lopez2016revisiting} introduced classifier two-sample tests (C2ST), where a binary classifier is trained to distinguish samples from two distributions, whereby the classifier's learned decision function and predictive uncertainty offer interpretive insight into distributional differences. However, these methods do not provide post-hoc explanations of existing two-sample tests, nor do they yield fine-grained attribution maps explaining which specific input features contribute most strongly to the test statistic, and thus to the statistical difference between the two groups, for individual samples.

\subsection{Contribution}
This work introduces a principled framework for explaining deep two-sample statistical tests. Our main contributions are: \textbf{(1) Problem formulation: explainability for deep statistical tests}. We formalize the problem of explaining deep two-sample tests by explicitly defining sample-level and feature-level contributions to population-level statistical significance. This establishes interpretability as an integral component of deep hypothesis testing, rather than a post-hoc analysis of model predictions.
\textbf{(2) Sample-level explainability of statistical significance}.
We introduce a sample-level influence score that measures how much each individual observation contributes to the statistical significance of the difference between two populations. By operating directly on the two-sample test statistic, the proposed score identifies samples that \textit{amplify} or \textit{attenuate} the rejection of the null hypothesis in a fully label-free manner.
\textbf{(3) Feature-level explainability of population differences}.
We derive a gradient-based feature attribution method that explains population-level statistical significance by backpropagating the two-sample test statistic through a deep representation. This yields spatially resolved feature importance maps that identify which regions of complex inputs drive the detected distributional differences between populations.
\textbf{(4) Empirical validation in biomedical imaging}.
We demonstrate the practical utility of our framework on T1 structural MRI data and fundus images of diabetic retinopathy, where the resulting explanations highlight biologically meaningful regions and samples associated with known disease markers, illustrating the interpretability of the proposed approach.
\section{Background}
\label{sec:format}
In this section, we provide a brief background on the two-sample tests used in this work. 
\subsection{Deep Two-Sample Test}
\label{subsec:two-sam}
We consider the non-parametric two-sample statistical testing, that is, determining whether two sets of observations are drawn from the same underlying distribution (the null hypothesis, denoted by $H_{0}: p=q$) or the samples are drawn from different distributions (the alternative hypothesis $H_{1}: p\neq q$) \cite{2sample-test}. Following \cite{2sample-test}, given two sets of observations $\mathcal{X}_n = \{x_i\}_{i=1}^n \sim p$ and $\mathcal{Y}_m = \{y_j\}_{j=1}^m \sim q$, a deep two-sample test first maps samples into a learned representation space using a neural network $\phi_{\theta}:\mathcal{X} \rightarrow \mathbb{R}^{H}$. The test then performs a multivariate location test on whether both means map to the same location. If the distance between the two means is too large, we reject the hypothesis that both samples are
drawn from the same distribution. Specifically, the test computes a test statistic based on the difference between the empirical means of the two populations in the representation space. The resulting test statistic, named Deep Maximum Mean Discrepancy (DMMD), takes the form
\begin{equation}
\label{eq:test-statistic}
    S(\mathcal{X}_{n}, \mathcal{Y}_{m}) = \frac{nm}{n+m} 
    \left\| \mu_{X} - \mu_{Y} \right\|^2 .
\end{equation}
Where \[
\mu_X = \frac{1}{n} \sum_{i=1}^n \phi_\theta(x_i), \quad
\mu_Y = \frac{1}{m} \sum_{j=1}^m \phi_\theta(y_j)
\] indicate the mean of embeddings on $\mathcal{X}_{n}$, and $\mathcal{Y}_{m}$ respectively. Given the test statistic $S{(\mathcal{X}_{n}, \mathcal{Y}_{m})}$, statistical significance is assessed under the null hypothesis $H_{0}$. As shown in \cite{2sample-test}, the null distribution of $S(\mathcal{X}_{n}, \mathcal{Y}_{m})$ can be approximated by a weighted sum of independent $\chi^{2}$ random variables. While this permits analytic p-value computation, evaluating the resulting distribution can be computationally expensive in practice. Instead, we estimate the null distribution using a Monte Carlo permutation test \cite{MCS}. Specifically, the learned representations are computed once for all samples, after which group labels are randomly permuted to generate empirical realizations of the test statistic under the null hypothesis. The p-value is then computed as the proportion of permuted test statistics exceeding the observed value. This resampling-based procedure is computationally efficient and is used throughout this work. This approach retains the power of deep representations while controlling the type-I error and achieving asymptotic consistency as sample size increases \cite{2sample-test}.
\section{Methodology}
\label{sec:methodology}
In this section, we present our approach for sample-level and feature-level explainability. Given two populations that are statistically different, our objective is to identify which specific samples or features contribute most to the observed significant differences between the groups. 
\subsection{Sample-Level Explainability}
\label{sec:sample_level}
We introduce a sample-level explainability framework that quantifies how each individual sample 
contributes to the statistical difference between two populations. 
\begin{definition}
\label{infscr}
The influence score of a sample $x_{i}$, denoted as $\mathcal{IF}(x_{i})$, is defined as the change in the test statistic when this sample is removed from $\mathcal{X}_{n}$:

\begin{equation}
    \mathcal{IF}(x_i) = S(\mathcal{X}_n, \mathcal{Y}_m) - S(\mathcal{X}_{n \setminus i}, \mathcal{Y}_m),
\end{equation}
where $\mathcal{X}_{n \setminus i} = \mathcal{X}_n \setminus \{x_i\}$ 
denotes the dataset with $x_i$ excluded.
Similarly, for a sample $y_j$ from $\mathcal{Y}_m$, the influence score is defined as:
\begin{equation}
    \mathcal{IF}(y_j) = S(\mathcal{X}_n, \mathcal{Y}_m) - S(\mathcal{X}_n, \mathcal{Y}_{m \setminus j}).
\end{equation}
\end{definition}
Intuitively, if $\mathcal{IF}(x_i) > 0$, the test statistic decreases when removing $x_i$, 
indicating that $x_i$ \textit{amplifies} the difference between groups—i.e., it contributes to the rejection of the null hypothesis and represents a distinctive sample.
Conversely, if $\mathcal{IF}(x_i) < 0$, removing $x_i$ increases the similarity between groups, suggesting that it \textit{suppresses} the detected difference. 
Samples with large positive influence scores are thus \textit{important} for the statistical significance 
of the group difference, while those with large negative scores indicate contribute to similarity  of two groups, indicating outliers or bridge samples between the two distributions.

\subsection{Feature-Level Explainability}
\label{Sec:FLE}
While sample-level explainability identifies which individual observations contribute most to statistical significance, feature-level explainability aims to localize where in each input the population-level differences arise. Our goal is to identify spatial regions whose representations contribute most strongly to the two-sample test statistic. We now describe our feature-level explainability framework for identifying the spatial regions that drive statistical differences between two populations. 
\begin{figure}[t]
  \centering
  \includegraphics[width=0.9\linewidth]{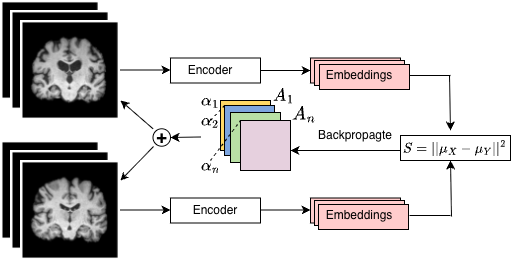}
  \caption{Feature-level explanation of population-level statistical significance. Two populations are encoded to embeddings to compute a two-sample statistic $S$, whose gradients are backpropagated to feature maps $A_{k}$ and aggregated via weights $\alpha_{k}$ to produce a spatial attribution map, which is then upsampled and overlaid on the input for visualization.}
  \label{fig:feature_level}
\end{figure}
\begin{definition}
Let $S(\mathcal{X}_{n}, \mathcal{Y}_{m})$ be the deep two-sample test statistic and let $\phi_{\theta}$ be the pretrained feature extractor used to compute the latent representations. For an input sample $x$, we define its feature-level statistical importance as the sensitivity of the test statistic with respect to spatially localized feature activations.
\end{definition}

Specifically, for two given populations $\mathcal{X}_{n}$ and $\mathcal{Y}_{m}$ we compute the  the test statistic $S(\mathcal{X}_{n}, \mathcal{Y}_{m})$. Then for each sample $x$ we 
backpropagate $S$ through the encoder to the last convolutional feature maps $A\in \mathbb{R}^{C\times h\times w}$ to assess the contribution of the internal features to the observed group-level difference.

In particular, let
$A_k \in \mathbb{R}^{h \times w}$ denote the $k^{th}$ feature map of the final convolutional layer. The gradient
$
\frac{\partial S}{\partial A_k^{pq}}
$
captures the sensitivity of the test statistic to activations at spatial location $pq$ within channel $k$. These gradients thus describe how variations in local image features influence the group discrepancy. 

To aggregate these local sensitivities into a spatial importance map, we compute channel importance weights $\alpha_{k}$, by computing the global average pooling (GAP):
\[
\alpha_k = \mathrm{GAP}\!\left(\frac{\partial S}{\partial A_k}\right)
= \frac{1}{hw} \sum_{p=1}^{h} \sum_{q=1}^{w}
\frac{\partial S}{\partial A_k^{pq}}.
\]
The final attribution map is obtained by a weighted linear combination of the
feature maps,
\[
M(x) = \mathrm{ReLU}\!\left(\sum_{k=1}^{C} \alpha_k A_k(x) \right),
\]
which highlights the spatial regions of the input that most strongly contribute to the statistical difference between the two groups. This construction yields a
feature-level explanation grounded in the sensitivity of the inter-group discrepancy to localized representation patterns. Our method is illustrated in Figure \ref{fig:feature_level}.  
While inspired by gradient-based attribution methods such as Grad-CAM \cite{gradcam}; our approach differs fundamentally in that it explains a population-level statistical significance rather than the prediction of an individual classifier.
\begin{figure*}[t]
\centering
\begin{minipage}[t]{0.33\textwidth}
  \centering
  \includegraphics[width=\linewidth]{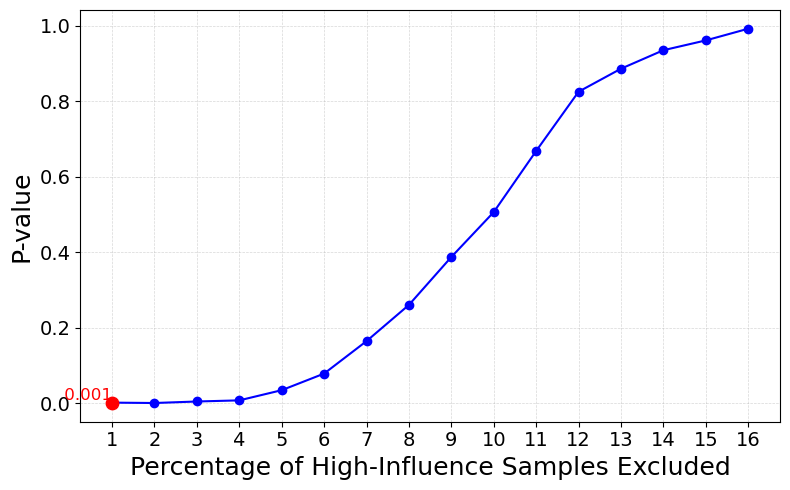}
   \centerline{(a) High-Influence Removal Effect}\medskip
 \label{fig_2_a} 
\end{minipage}
\hfill
\begin{minipage}[t]{0.33\textwidth}
  \centering
  \includegraphics[width=\linewidth]{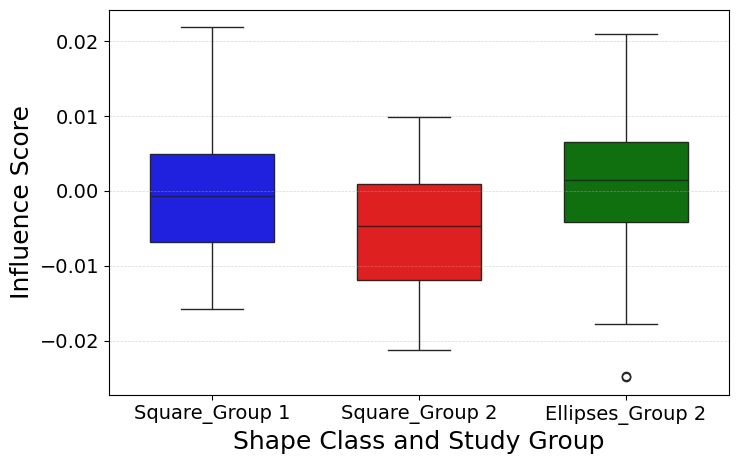}
   \centerline{(b) Distribution of Influence Scores}\medskip
  \label{fig_2_b}
\end{minipage}
\hfill
\begin{minipage}[t]{0.32\textwidth}
  \centering
  \includegraphics[width=\linewidth]{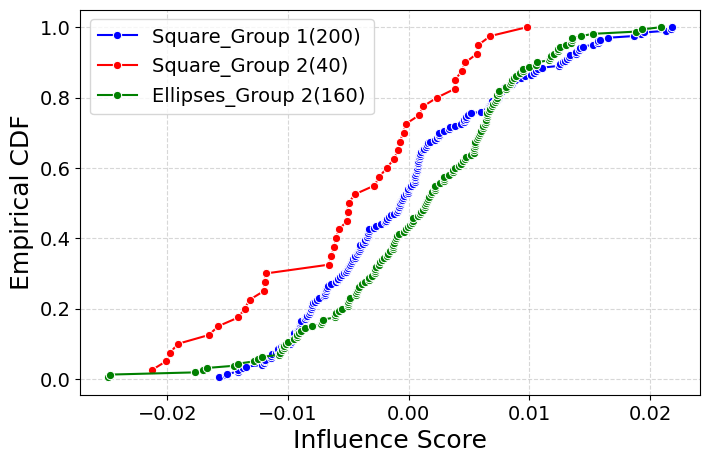}
   \centerline{(c) CDF of Infleuence Scores}\medskip
  \label{fig_2_c}
\end{minipage}
\hfill
\caption{Sample-level explainability results on dSprites.}
\label{SL-DS}
\end{figure*}
\vspace{-0.3em}
\begin{figure*}[t]
\centering
\begin{minipage}[t]{0.33\textwidth}
  \centering
  \includegraphics[width=\linewidth]{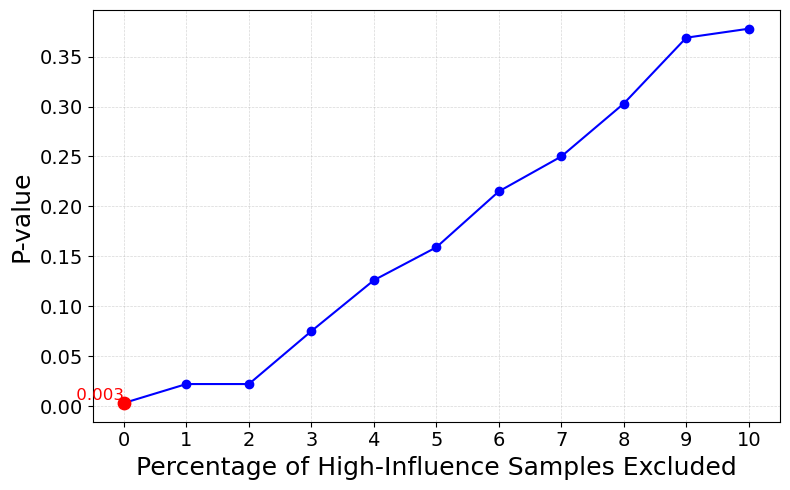}
   \centerline{(a) High-Influence Removal Effect}\medskip
 \label{fig_d_a} 
\end{minipage}
\hfill
\begin{minipage}[t]{0.33\textwidth}
  \centering
  \includegraphics[width=\linewidth]{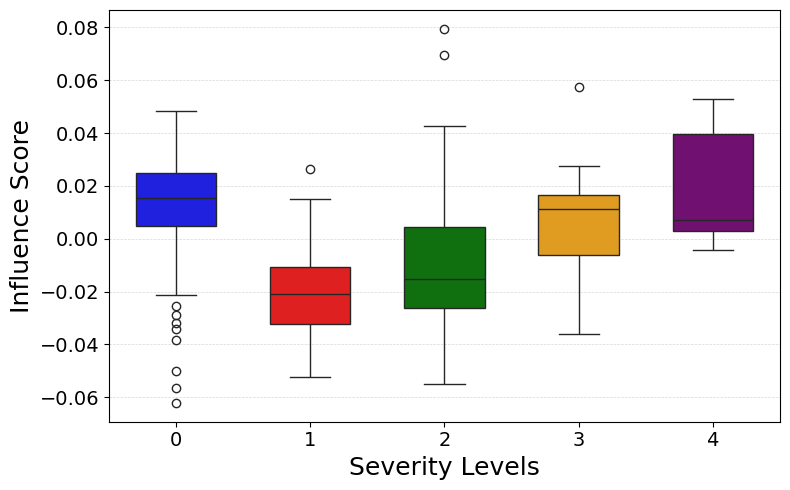}
   \centerline{(b) Distribution of Influence Scores}\medskip
  \label{fig_d_b}
\end{minipage}
\hfill
\begin{minipage}[t]{0.32\textwidth}
  \centering
  \includegraphics[width=\linewidth]{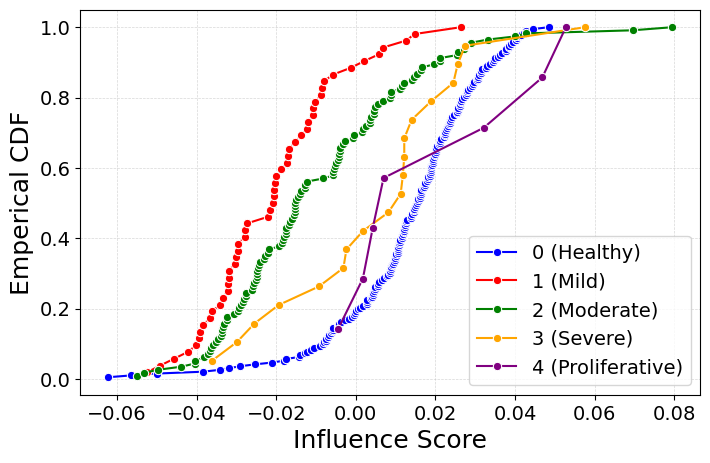}
   \centerline{(c) CDF of Influence Scores}\medskip
  \label{fig_d_c}
\end{minipage}
\hfill
\caption{Sample-level explainability results on Diabetic Retinopathy.}
\label{SL-DR}
\end{figure*}
\section{Experiments}
\label{sec:Experiments}
In this section, we conduct qualitative and quantitative evaluations of the proposed sample-level and feature-level explainability methods for the deep two-sample test. We first perform controlled experiments on synthetic and simulated data to verify that the proposed explanations are consistent with known distributional differences. We then demonstrate the practical utility of both sample- and feature-level explanations on real biomedical imaging datasets, including structural MRI and diabetic retinopathy fundus images, where the resulting explanations provide insight into disease-related population differences. Finally, we illustrate additional applications of the proposed pipeline.
\subsection{Setup}
We evaluate the proposed explainability framework on three datasets with corresponding feature extractors. Specifically, we consider one synthetic dataset, dSprites \cite{dSpreites}, and two biomedical imaging datasets: T1-weighted structural MRI data from the Alzheimer’s Disease Neuroimaging Initiative (ADNI) \cite{ADNI} and retinal fundus images from the EyePACS Diabetic Retinopathy Detection dataset \cite{EyePACS}. 
For the dSprites dataset, we train a variational autoencoder (VAE) on the heart-shaped subset and use the encoder to obtain latent representations. This yields 10-dimensional latent representations. For T1-weighted structural MRI, we use a ResNet-50 encoder trained in a supervised manner on the UK Biobank dataset to predict chronological age; the pretrained encoder is then used to extract embeddings for the ADNI subjects. For the diabetic retinopathy experiments, we employ a self-supervised SimCLR model \cite{simclr} with a ResNet-50 backbone trained using contrastive learning on the APTOS dataset \cite{APTOS}. The resulting encoder is used to compute embeddings for retinal fundus images in the EyePACS dataset. The ResNet-50 backbone produces 2048-dimensional representations. None of the encoders is fine-tuned on the test datasets used for two-sample testing. To assess statistical significance between the two groups, we follow the procedure described in Section \ref{subsec:two-sam}, with additional details provided in Supplementary Section 1. 
\begin{figure*}[t]
\centering
\begin{minipage}[t]{0.45\textwidth}
  \centering
  \textbf{High-Influence Samples}\\[0.2em]
  \includegraphics[width=\linewidth]{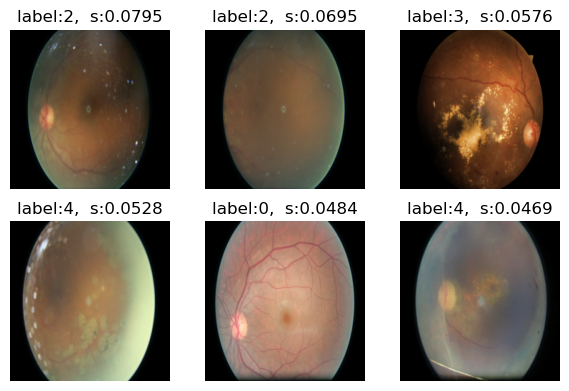}
\end{minipage}
\hspace{1.5em}
\begin{minipage}[t]{0.45\textwidth}
  \centering
  \textbf{Low-Influence Samples}\\[0.2em]
  \includegraphics[width=\linewidth]{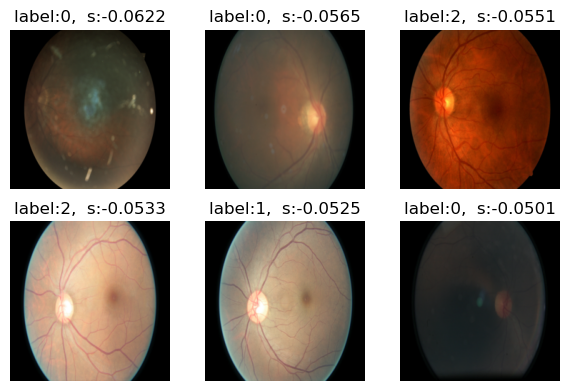}
\end{minipage}
\caption{
Qualitative examples illustrating samples with high (left) and low (right) influence scores.
High-influence samples exhibit clear and distinctive visual patterns that strongly contribute
to group differences, while low-influence samples appear more ambiguous or borderline.
}
\label{fig:QA}
\end{figure*}
\subsection{Sample-Level Explainability}
\label{subsec:samlevel}
We evaluate our method in two complementary settings. First, we perform controlled experiments on the synthetic dSprites dataset, which provides known ground-truth factors of variation. This setting serves as a sanity check to validate the expected behavior of influence scores under controlled conditions. Second, we consider a real-world biomedical imaging setting using retinal fundus images from the EyePACS Diabetic Retinopathy dataset, where group separation arises from varying disease severity. This setting assesses whether influence scores identify samples and features associated with clinically meaningful patterns. 
\subsubsection{dSprites} 
We consider two groups from the dSprites dataset, each group containing 200 samples. The first group $\mathcal{X}$ contains images with square shapes only, while the second, $\mathcal{Y}$ contains images with both squares ($\# 40$) and ellipses ($\# 160$). Ellipses are the sole factor inducing group separation, whereas squares are identical across both groups and encode similarity. We use the encoder of trained VAE on the heart class for getting embeddings in each group and compute DMMD $S(\mathcal{X}, \mathcal{Y})$ based on Equation \ref{eq:test-statistic}. Consequently, we compute $\mathcal{IF}(x)$ for each sample $x$, following Definition \ref{infscr}.    
\paragraph{Evaluation Procedure.} To evaluate whether our method correctly identifies the important samples for the statistical difference between two groups, we monitor the p-value. If the proposed influence score correctly identifies the samples that contribute positively to the
distinction between the two groups, excluding these samples should result in an increase in the p-value (i.e., by excluding them, the two groups should become more similar). To assess this, we rank the samples by influence scores $\mathcal{IF}(x)$ and progressively remove the
highest ones to analyze their impact on the p-value. We then recompute the test statistic and p-value and plot the results as a function of the proportion of removed samples.
\paragraph{Results.} 
The results of the simulated anaylsis on the dSprites dataset are summarized in Figure \ref{SL-DS}.
First, we monitor the p-value as samples with the highest $\mathcal{IF}$ scores are progressively removed. As shown in Figure \ref{SL-DS} (a), the p-value increases as more high-influence samples are excluded, indicating that these samples are responsible for amplifying the statistical difference between the two groups. 
To further characterize the distribution of influence scores, we report both box plots and cumulative distribution functions (CDFs) for samples in each group. The box plots in Figure \ref{SL-DS} (b) indicate that ellipse samples have a median influence score above zero, demonstrating their positive contribution to group separation, whereas square samples in Group 2 exhibit median influence scores below zero, consistent with their role in similarity between groups. This behavior is further corroborated by the CDFs in Figure \ref{SL-DS} (c), where the influence-score distribution for ellipses is right-shifted, while the distribution for square samples in Group 2 is left-shifted. Together, these results confirm that the proposed influence scores correctly identify samples that drive statistical differences as well as samples that promote similarity between the two groups.

\subsubsection{Diabetic Retinopathy} 
We consider two groups of retina images, healthy and unhealthy images, each containing 200 samples. The healthy group consists of retina images with no signs of diabetic retinopathy. The unhealthy group includes images with varying severity levels of diabetic retinopathy, ranging from 1 to 4. To ensure a representative distribution of severity levels within the unhealthy group, we use stratified sampling, preserving the original proportions of each severity level. We use the encoder of SimCLR model pre-trained on APTOS dataset, to get embeddings in each group and compute the DMMD test statistic and influence score $\mathcal{IF}(x)$ for each sample $x$ following Definition \ref{infscr}. 

\begin{figure}[t]
\centering
\begin{minipage}[t]{0.23\linewidth}
  \centering
  \includegraphics[width=\linewidth]{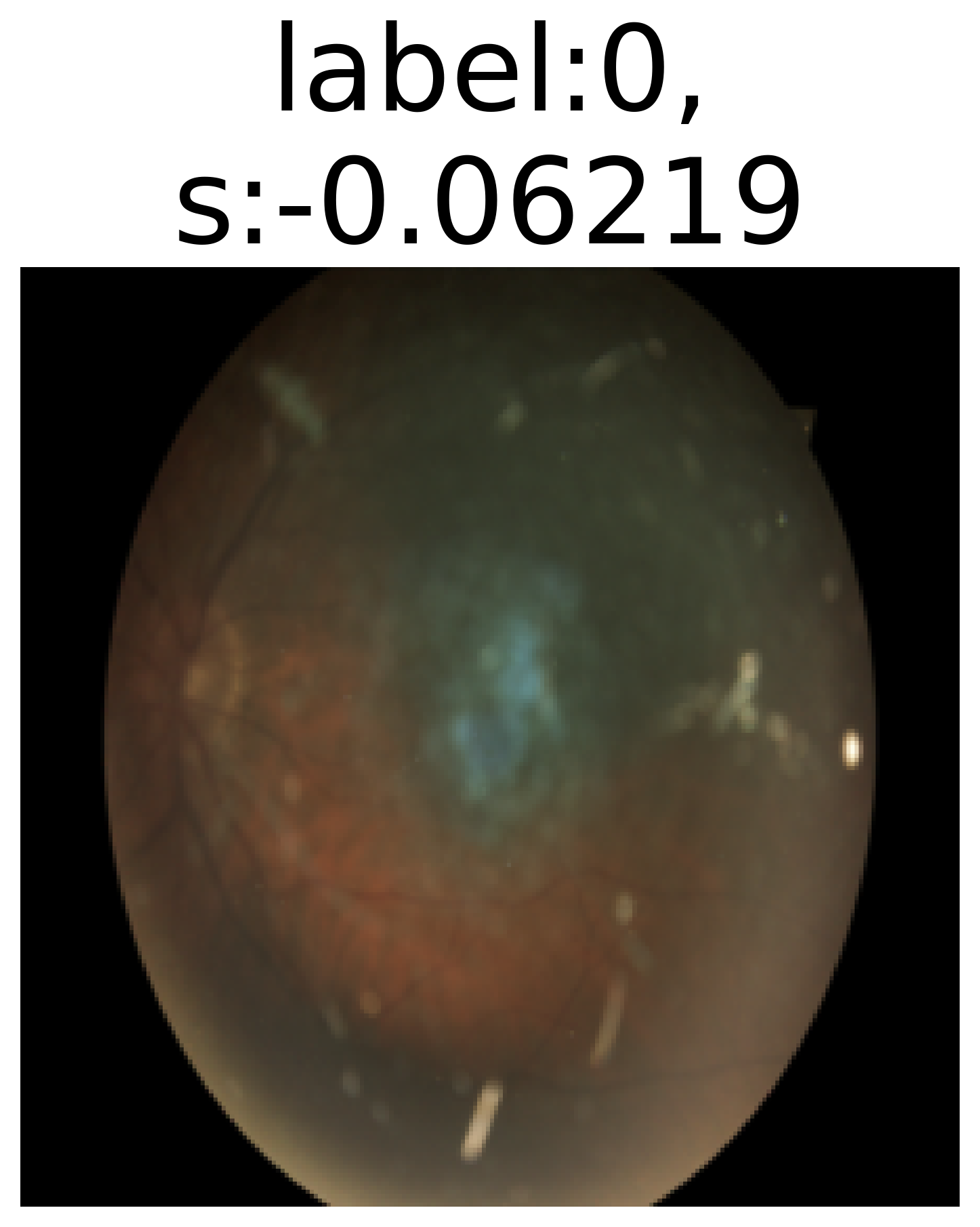}\\
  \small (0)
\end{minipage}
\hfill
\begin{minipage}[t]{0.23\linewidth}
  \centering
  \includegraphics[width=\linewidth]{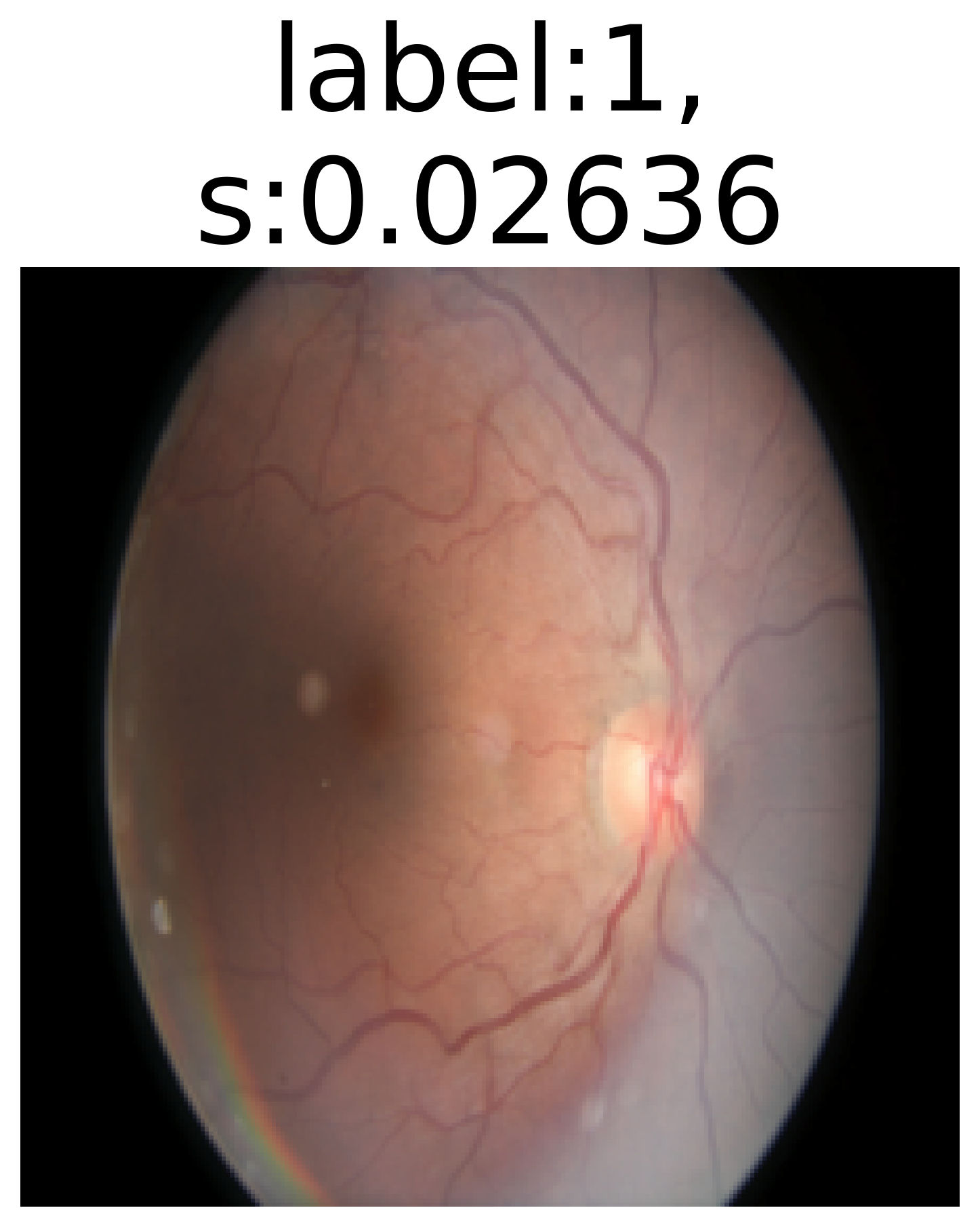}\\
  \small (1)
\end{minipage}
\hfill
\begin{minipage}[t]{0.23\linewidth}
  \centering
  \includegraphics[width=\linewidth]{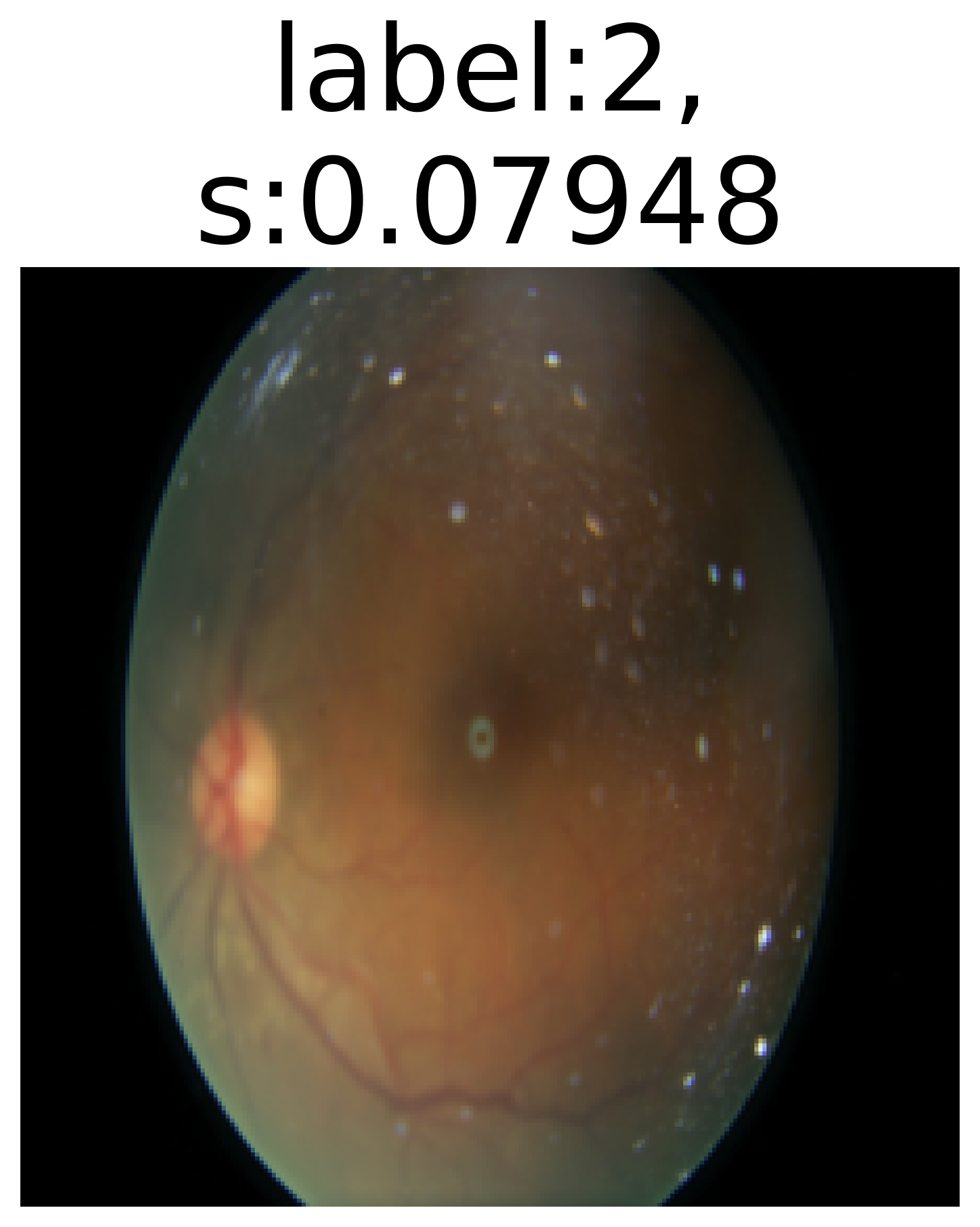}\\
  \small (2)
\end{minipage}
\hfill
\begin{minipage}[t]{0.23\linewidth}
  \centering
  \includegraphics[width=\linewidth]{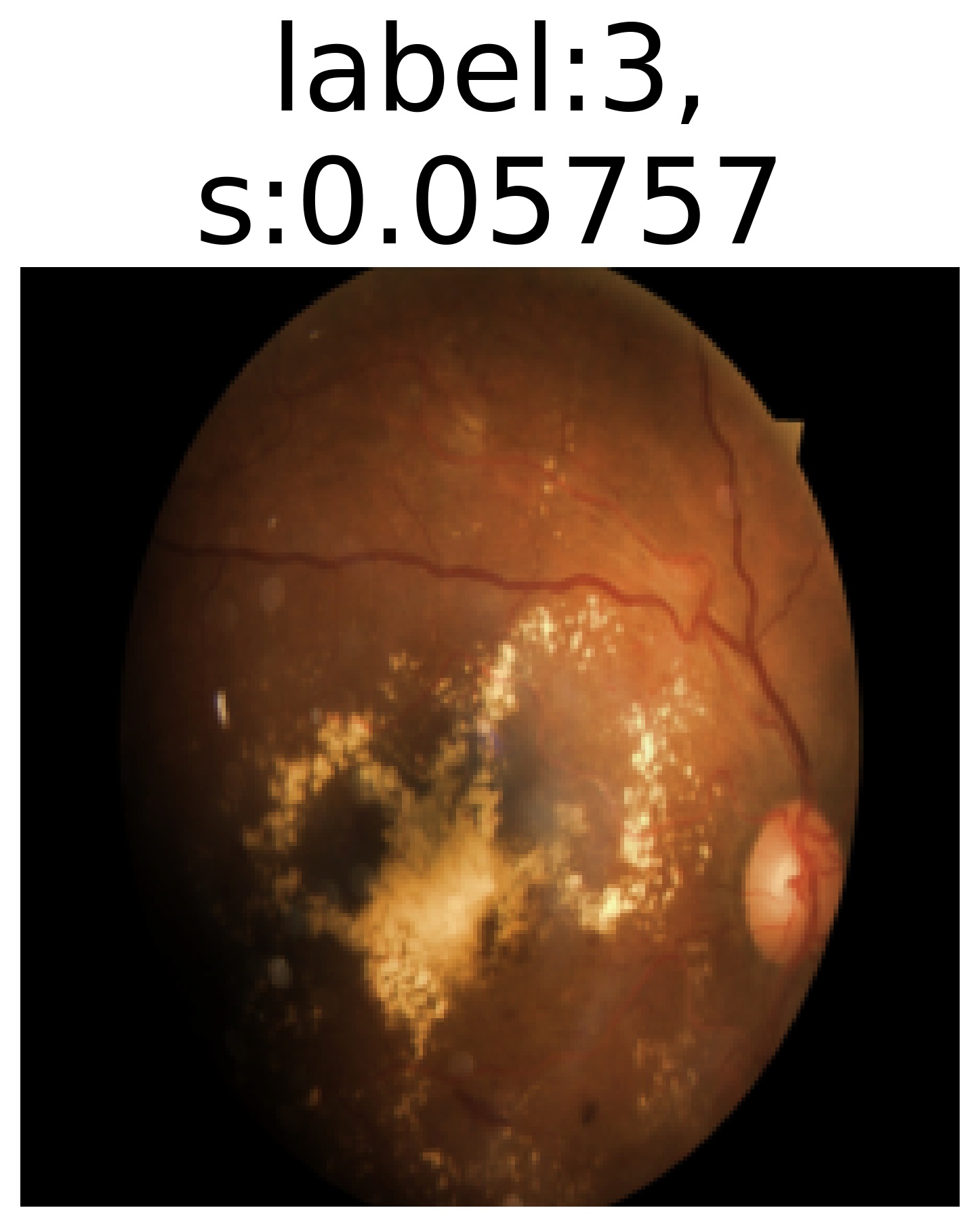}\\
  \small (3)
\end{minipage}
\caption{
Representative outliers identified by influence scores for diabetic retinopathy severity levels 0–3.
No outliers were observed for severity level 4 under the chosen criterion.
}
\label{fig:out}
\end{figure}

\paragraph{Quantitative Results.} Figure \ref{SL-DR} summarizes quantitative results for diabetic retinopathy dataset. As shown in Figure \ref{SL-DR} (a), similar to dSprites dataset, the p-value increases as samples with high influence scores are progressively removed. In contrast, excluding samples with low influence scores leads to a decrease in the p-value (see Supplementary Figure 1), confirming that these samples contribute to the similarity between the two groups rather than their separation. We further analyze the distribution of influence scores across disease severity levels in Figure \ref{SL-DR} (b). The influence scores tend to increase with severity, suggesting that more severe cases contribute more strongly to group differences. Moreover, this trend is monotonic, confirming that our method captures clinically meaningful variation. The plot also reveals the presence of outliers within each severity group, which correspond to samples whose appearance deviates from their clinical labels, potentially due to early-stage disease, atypical presentations, or image-quality variability. The CDF plot in Figure \ref{SL-DR} further supports these findings. Healthy images and severe disease cases exhibit positive median influence scores, whereas mild disease cases often show negative influence scores, reflecting their visual similarity to healthy images. 
\paragraph{Qualitative Results.} Figure \ref{fig:QA} presents qualitative examples of high- and low-influence samples. High-influence images exhibit strong, unambiguous visual patterns that clearly distinguish healthy from unhealthy cases—such as pronounced lesions, clear pathological structures, or very typical healthy appearance. While low-influence samples appear ambiguous, borderline, or visually similar to the opposite group. Together, these examples show that influence scores align well with visual interpretability and capture clinically meaningful variability.
Figure \ref{fig:out} further visualizes representative outliers from each group. These samples exhibit atypical appearance, borderline severity, or reduced image quality, highlighting that influence scores surface interpretable edge cases and heterogeneity rather than arbitrary noise.

\subsection{Feature-Level Explainability}
\label{subsec:Feat_level}
We evaluate feature-level explanations in two complementary settings that reflect different levels of supervision. First, we consider the synthetic dSprites dataset, where ground-truth generative factors are known. This setting allows us to directly assess whether the proposed feature-level explanations correctly identify the latent factors responsible for group differences. Second, we evaluate feature-level explanations on real neuroimaging from the ADNI dataset, where ground-truth feature importance is unavailable. In this case, we assess whether the identified features align with established clinical and neuroanatomical knowledge, thereby evaluating the clinical plausibility of the explanations.
\subsubsection{dSprites} 
\begin{figure}[t]
\centering
\setlength{\tabcolsep}{4pt}
\begin{tabular}{lclc}
Square &  & Ellipse &  \\
\includegraphics[width=0.18\linewidth]{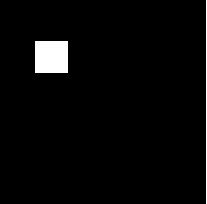} &
\includegraphics[width=0.18\linewidth]{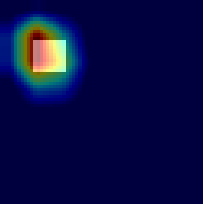} &
\includegraphics[width=0.18\linewidth]{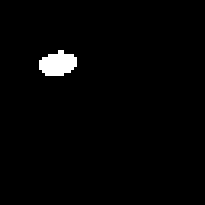} &
\includegraphics[width=0.18\linewidth]{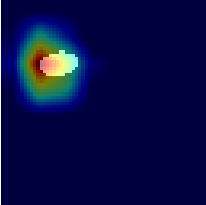} \\
\includegraphics[width=0.18\linewidth]{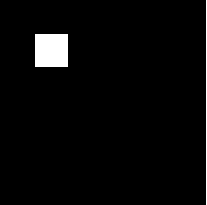} &
\includegraphics[width=0.18\linewidth]{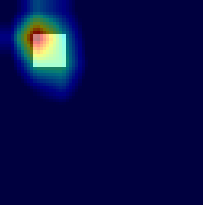} &
\includegraphics[width=0.18\linewidth]{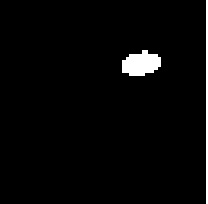} &
\includegraphics[width=0.18\linewidth]{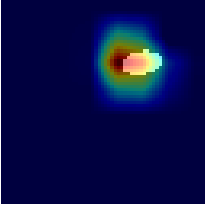}
\end{tabular}
\caption{Feature-level explanations on dSprites.}
\label{fig:flds}
\end{figure}

We consider two groups from the dSprites dataset, each group containing 78208 samples. The first group contains images with square shapes, while the second contains images with elliptical shapes. We use the encoder of pretrained VAE on the heart class to extract latent embeddings for each group and compute the DMMD statistic $S(\mathcal{X}, \mathcal{Y})$ based on Equation \ref{eq:test-statistic}. We then apply the proposed feature-level explanation method (Section \ref{Sec:FLE}) to obtain attribution heatmaps for individual samples. For dSprites, we generate explanations by backpropagating the test statistic to the penultimate convolutional layer rather than the final one, as the final layer has insufficient spatial resolution to produce meaningful heatmaps. This choice follows common practice in gradient-based attribution methods, which select the deepest layer with sufficient spatial resolution. For visualization, attribution maps are upsampled to the input resolution using bilinear interpolation and overlaid on the original images.
\paragraph{Results.} Figure \ref{fig:flds} presents representative qualitative results. The resulting heatmaps correctly highlight shape-specific regions (corner vs curvature), indicating that the proposed method accurately identifies the generative factors responsible for the observed group differences.
\subsubsection{ADNI} 
We consider two groups constructed from the ADNI dataset based on hippocampal volume, a well-established neuroimaging biomarker of Alzheimer’s disease–related neurodegeneration. Each group includes 918 2D slices. Subjects exhibiting reduced hippocampal volume indicative of hippocampal atrophy are assigned to Group 1, while subjects with preserved hippocampal volume form Group 2. This stratification induces group differences associated with disease-relevant anatomical variation and provides a clinically meaningful setting for evaluating whether the proposed feature-level explanations highlight brain regions known to be implicated in Alzheimer’s disease/cognitive impairment.

We use a ResNet-50 encoder pretrained on the UK Biobank dataset, predicting the age of brains, and adapt this pretrained model using a linear probing strategy on the ADNI dataset. Specifically, we train a logistic regression layer on top of the frozen encoder to predict group membership, while keeping all encoder parameters fixed. After training, the linear head is discarded, and the frozen encoder is used to extract embeddings for two-sample testing and for generating feature-level explanations. We apply the proposed feature-level explanation method to individual MRI scans and assess whether the resulting attributions align with known disease-related neuroanatomical patterns.

To assess the robustness of our feature-level explanations, we apply multiple channel-aggregation strategies to the gradients of the test statistic:
(i) gradient-weighted aggregation \cite{gradcam}, (ii) second-order gradient aggregation \cite{gradcampp}, and (iii) layer-wise spatial aggregation \cite{lcam}.
Across all variants, the resulting heatmaps remain consistent. Additional details are provided in the Supplementary Material.
\paragraph{Results.}
Figure \ref{fig:adni_heatmaps} indicates the qualitative results. Our explainability method consistently highlights the hippocampal and medial temporal lobe regions, and the resulting heatmaps show strong correspondence with areas affected by hippocampal atrophy. These results supports that the identified regions align with established AD biomarkers.

\begin{figure*}[t]
\centering
\setlength{\tabcolsep}{2pt}

\begin{tabular}{ccccc}
 & Original & Gradient-weighted & Second-order & \hspace{-1.6em} Layer-wise \\
\raisebox{.75\height}{Group 1} &
\includegraphics[width=0.15\textwidth]{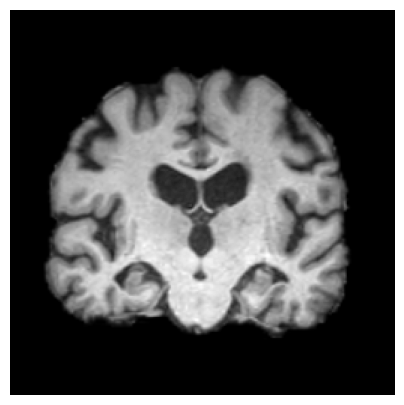} &
\includegraphics[width=0.15\textwidth]{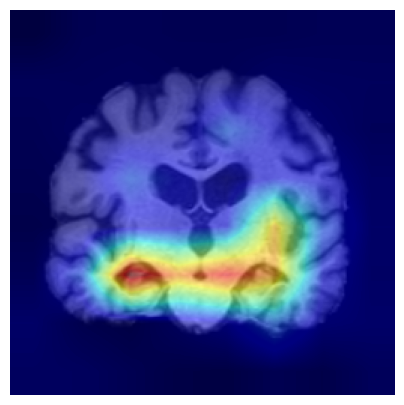} &
\includegraphics[width=0.15\textwidth]{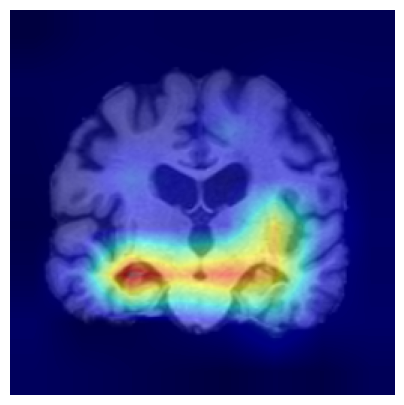} &
\includegraphics[width=0.175\textwidth]{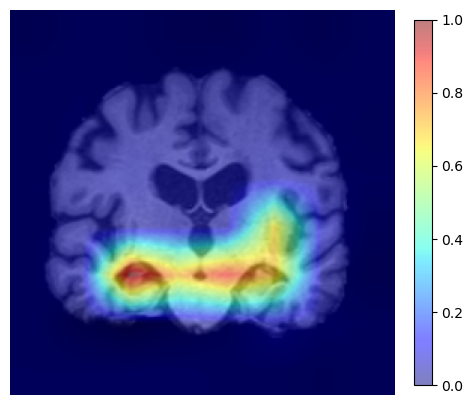} \\

\raisebox{.75\height}{Group 2} &
\includegraphics[width=0.15\textwidth]{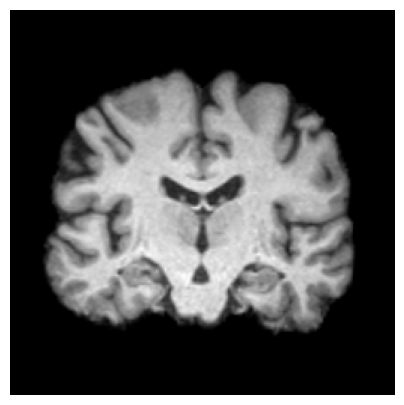} &
\includegraphics[width=0.15\textwidth]{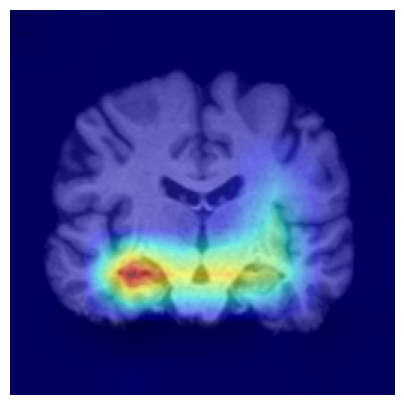} &
\includegraphics[width=0.15\textwidth]{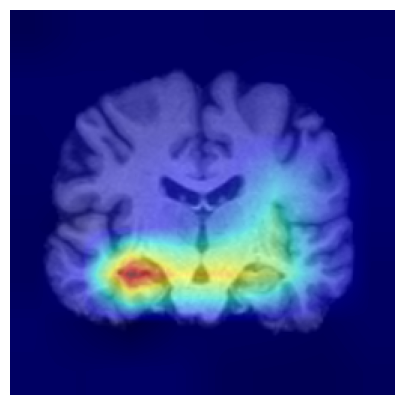} &
\includegraphics[width=0.175\textwidth]{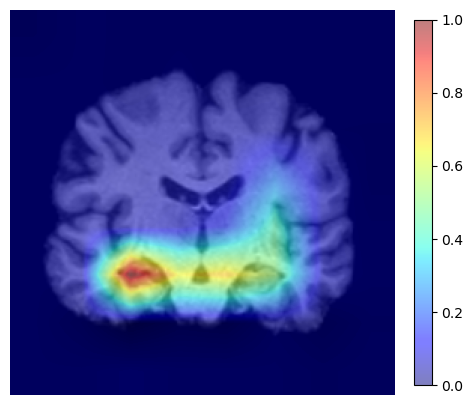} \\
\end{tabular}
\caption{\textbf{Feature-level explanations on ADNI.}
Top row: representative subject from Group 1. Bottom row: representative subject from Group 2.
Heatmaps are obtained by backpropagating the two-sample test statistic $S$ to the final convolutional layer.
Gradient-weighted, Second-order, and Layer-wise differ only in aggregation strategy but explain the same statistical objective.}
\label{fig:adni_heatmaps}
\end{figure*}

\section{Discussion}
\label{sec:dis}
We introduced an explainable deep statistical framework for identifying which samples and which features drive statistically significant group differences. We validated the framework on both synthetic data, where ground-truth generative factors are known, and on real biomedical data, where the resulting explanations align with established disease-related patterns. The proposed framework provides complementary insights at the sample level and the feature level, each of which supports distinct but related biomedical use cases. 
\subsection{Use Cases}
\paragraph{Sample-Level Explainability.} \textbf{1) Patient Stratification}: By identifying which samples drive the statistical separation between groups, our method enables the identification of patients with strong vs. weak disease signatures. This allows us to quantify heterogeneity within clinical cohorts and stratify patients based on biologically meaningful imaging patterns, which is particularly valuable in heterogeneous diseases. For example, in ADNI, clinically defined groups such as cognitively normal (CN), mild cognitive impairment (MCI), and Alzheimer’s disease (AD) are available. Using sample-level explainability, we can identify which individual subjects most strongly drive the separation between groups, e.g., in CN vs. MCI. High-influence MCI subjects exhibit imaging patterns that are more distinct from CN, potentially reflecting more aggressive or advanced disease processes. In contrast, low-influence MCI subjects appear more similar to CN, suggesting milder or earlier-stage pathology. By comparing the imaging features highlighted in high- vs. low-influence MCI subjects, we gain insights into heterogeneity within the MCI population and potential trajectories of disease progression. The same framework can be applied to MCI vs. AD to identify individuals transitioning toward more advanced pathology and to characterize imaging features associated with early progression.
\textbf{2) Trial Enrichment}: In a clinical trial setting, our sample-level explainability allows us to identify patients who contribute most strongly to the statistical separation between a disease group and a reference population. These high-influence patients typically exhibit clearer and more consistent pathological imaging patterns. By identifying patients who most strongly express the target pathology, we enrich trials with subjects more likely to show treatment effects. This reduces noise by excluding low-informative or ambiguous cases identified by low influence scores. This can improve statistical power and reduce the required sample size.
\textbf{3) Quality Control}: The method naturally supports quality control by flagging outlier samples with unexpected influence scores, detecting mislabeled, low-quality, or artifact-driven images, and supporting dataset curation prior to downstream modeling or clinical trials.
\paragraph{Feature-Level Explainability.} At the feature level, the method highlights spatial patterns that drive statistically significant group differences, helping identify candidate imaging biomarkers or disease-related patterns that are both discriminative and interpretable. These explanations can support biomarker discovery by identifying candidate regions or structures associated with disease-related variation. The framework also enables hypothesis-driven analyses: by defining groups based on clinically or biologically meaningful variables, researchers can use the resulting explanations to explore specific mechanisms, such as progression effects, sex differences, or subtype-related patterns. This supports hypothesis generation and provides interpretable insights that can guide downstream experimental validation.

Overall, the proposed method turns statistical significance into actionable insight, supporting patient stratification, quality control, trial enrichment, and biomarker discovery by making group differences explainable and biologically meaningful. The code is provided as a ZIP archive in the Supplementary Material.
\section*{Acknowledgments}
We gratefully acknowledge funding by the Deutsche Forschungsgemeinschaft (DFG, German Research Foundation) – project number 459422098.

\newpage

\bibliographystyle{named}
\bibliography{refs}

\end{document}